\documentclass[electronic]{vgtc}             

\usepackage{paralist}

\ifpdf
  \pdfoutput=1\relax                   
  \usepackage{graphicx}                
  \DeclareGraphicsExtensions{.pdf,.png,.jpg,.jpeg} 
\else
  \usepackage{graphicx}                
  \DeclareGraphicsExtensions{.eps}     
\fi%

\graphicspath{{figures/}{pictures/}{images/}{./}} 

\usepackage{microtype}                 
\PassOptionsToPackage{warn}{textcomp}  
\usepackage{textcomp}                  
\usepackage{mathptmx}                  
\usepackage{times}                     
\usepackage{cite}                      
\usepackage{tabu}                      
\usepackage{amsmath}                   
\usepackage{amssymb}                   

\usepackage[table,svgnames]{xcolor}
\usepackage{todonotes}
\usepackage{amsfonts}
\usepackage{tabularx}

\onlineid{0}

\vgtccategory{Research}

\title{
``Normalized Stress" is Not Normalized: How to Interpret Stress Correctly
}

\author{
Kiran Smelser\thanks{e-mail: ksmelser@arizona.edu},\hspace{0.5cm}  %
Jacob Miller\thanks{e-mail: jacobmiller1@arizona.edu}, \hspace{0.5cm} %
Stephen Kobourov\thanks{e-mail: kobourov@cs.arizona.edu}
}
\affiliation{\scriptsize University of Arizona  }

\teaser{
 \centering
 \includegraphics[width=.85\linewidth]{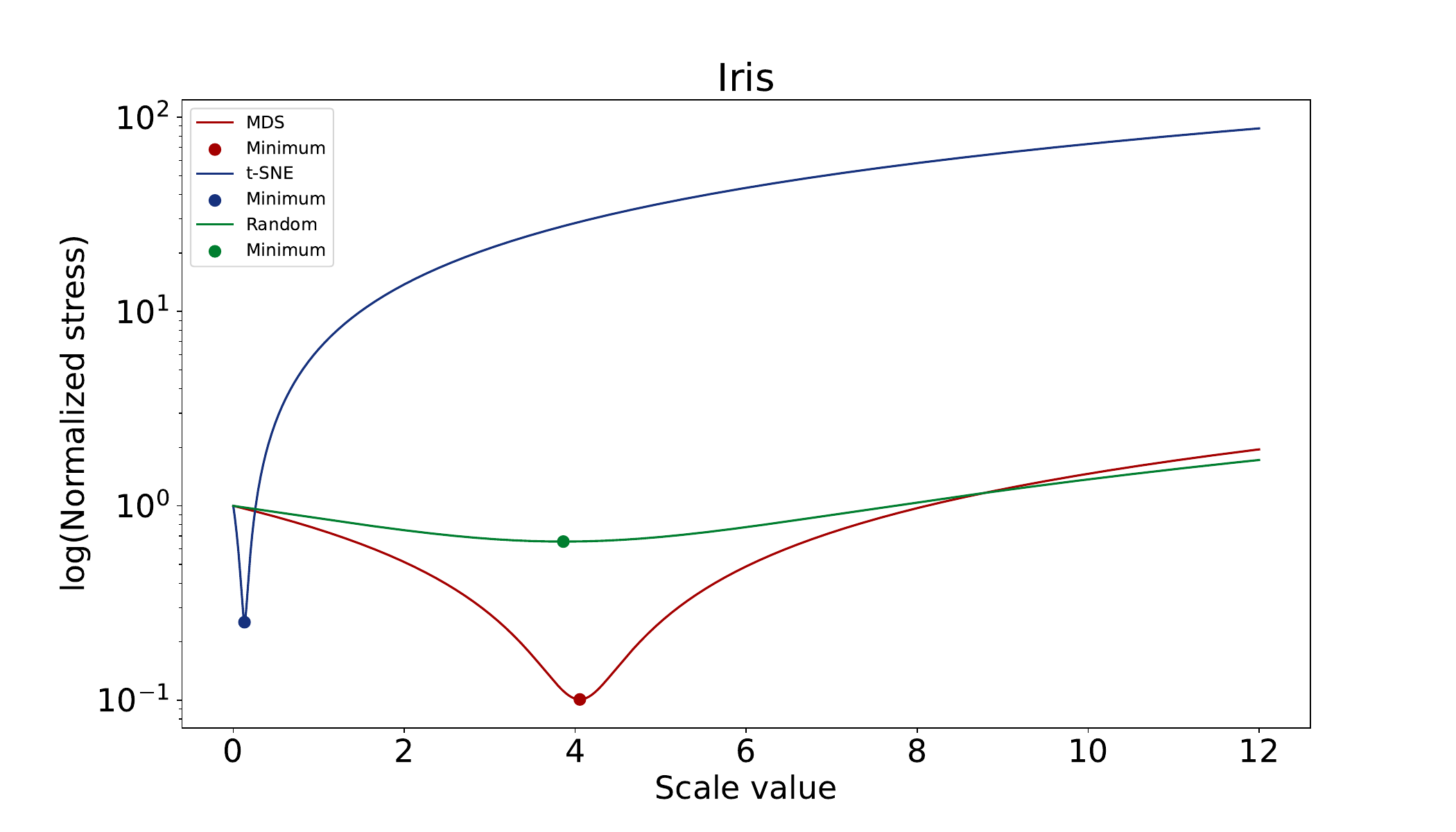}

 \includegraphics[width=0.32\linewidth]{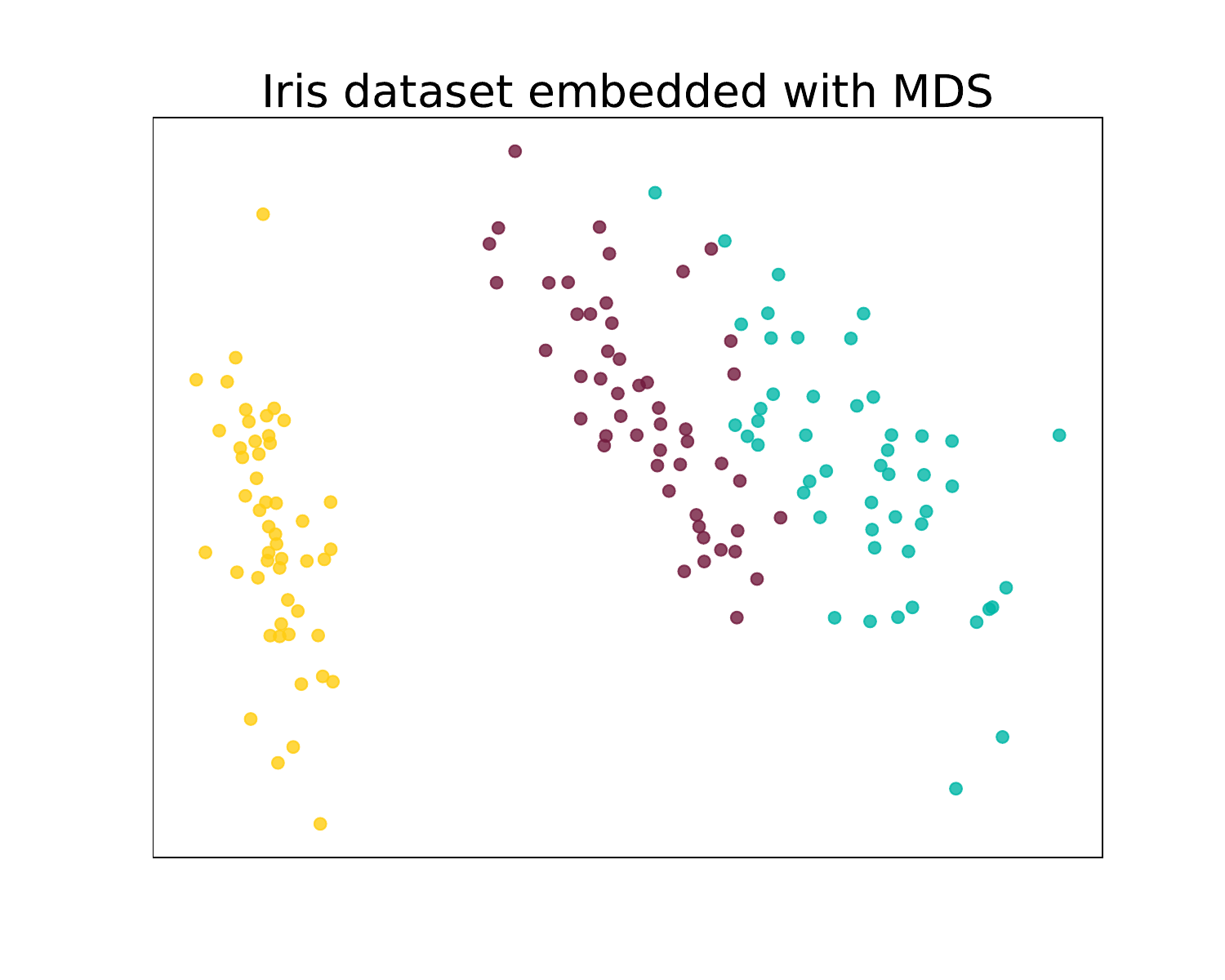}
 \includegraphics[width=0.32\linewidth]{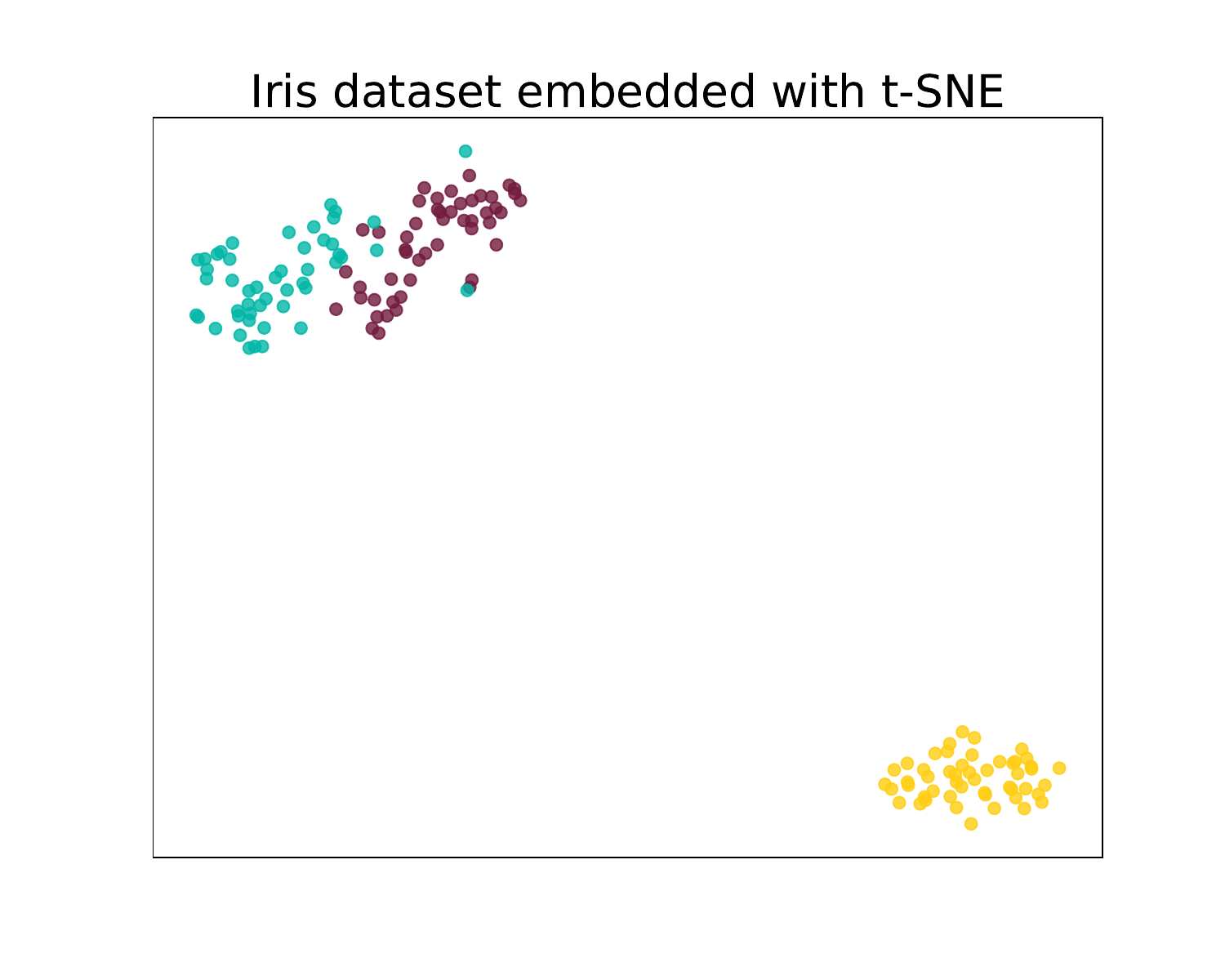}
 \includegraphics[width=0.32\linewidth]{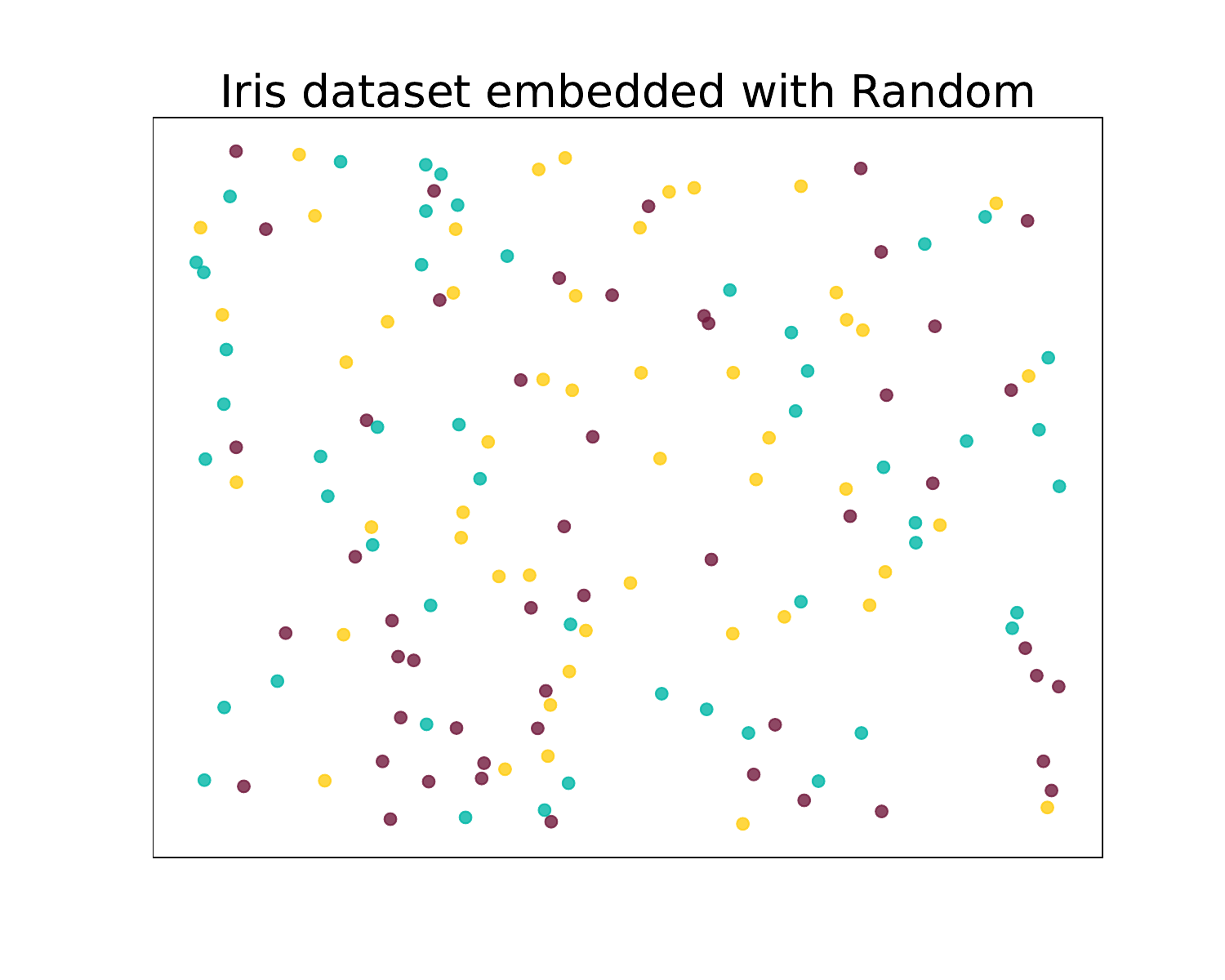} 
\caption{
{MDS~\cite{shepard1962analysis}, t-SNE~\cite{van2008visualizing}, and RND (random) embeddings of the well-known $Iris$ dataset from left to right (bottom).}
The plot (top) shows the values of the {\em normalized stress} metric for these three embeddings and clearly illustrates the sensitivity to scale. As one uniformly scales the embeddings to be larger or smaller, the value of normalized stress changes. Notably, at different scales, different embeddings have lower stress, including the absurd situation where the random embedding has the lowest stress (beyond scale 9). 
Moreover, the expected order of MDS$<$t-SNE$<$RND is only found briefly at a scalar value slightly greater than 0.25 (hardly visible in the plot), and all six different algorithm orders can be found by selecting different scales.}
 \label{fig:teaser}
}

\abstract{
Stress is among the most commonly employed quality metrics and optimization criteria for dimension reduction projections of high-dimensional data. Complex, high-dimensional data is ubiquitous across many scientific disciplines, including machine learning, biology, and the social sciences. One of the primary methods of visualizing these datasets is with two-dimensional scatter plots that visually capture some properties of the data. 
Because visually determining the accuracy of these plots is challenging, researchers often use quality metrics to measure the projection’s accuracy or faithfulness to the full data.
One of the most commonly employed metrics, normalized stress, is sensitive to uniform scaling (stretching, shrinking) of the projection, despite this act not
meaningfully changing anything about the projection. We investigate the effect of scaling on stress and other distance-based quality metrics analytically and empirically by showing just how much the values change and how this affects dimension reduction technique evaluations. We introduce a simple technique to make normalized stress scale-invariant and show that it accurately captures expected behavior on a small benchmark. 
} 

\CCScatlist{ 
    Dimension reduction, 
    Empirical evaluation, Stress
}

\begin{document}


\firstsection{Introduction}

\maketitle


Dimensionality reduction (DR) techniques play an integral
role in the visualization of complex, high-dimensional datasets~\cite{10.5555/383784.383790,DBLP:conf/vissym/LiuMWBP15,DBLP:journals/tvcg/KehrerH13,DBLP:conf/www/TangLZM16}. DR techniques, which map high-dimensional data to the 2D computer screen, are pivotal in various fields, from bioinformatics~\cite{huang2022towards,DBLP:journals/bib/MaD11,DBLP:journals/bib/HilarioK08} to machine learning (ML)~\cite{DBLP:journals/corr/MakhzaniSJG15,DBLP:journals/corr/abs-2103-11251,osti_15002155}. 

Researchers typically evaluate the effectiveness of DR techniques such as Multidimensional Scaling (MDS)~\cite{torgerson1952multidimensional,shepard1962analysis,kruskal1964multidimensional} and t-Distributed Stochastic Neighbor Embedding (t-SNE)~\cite{van2008visualizing} using quantitative quality metrics.
Many of 
these metrics measure how well an embedding technique performs in preserving the structure and relationships inherent in the original high-dimensional data when projected onto a lower-dimensional space~\cite{DBLP:journals/tvcg/EspadotoMKHT21,DBLP:journals/tvcg/NonatoA19,DBLP:journals/ijon/LeeV09,DBLP:journals/ijon/MokbelLGH13,georg2004survey}.

One aspect that has received little attention is 
the effect that scaling the projected data has on these quality metrics which, to our knowledge, has not been well studied in the visualization literature. 
{
~\autoref{fig:teaser} shows a simple example of how much scale can affect evaluation results of stress -- by merely resizing the outputs of different techniques, one can dramatically alter the evaluation outcome, often leading to absurd conclusions (such as a random embedding having the lowest stress).
It should be clear that normalized stress (perhaps often confused with the scale-invariant non-metric stress of Kruskal~\cite{kruskal1964multidimensional}) is sensitive to scale.
}

Comparing embeddings generated by various DR techniques using normalized stress {\em without taking scale into account} can be seen as picking an arbitrary point on the stress-scale curve for \textit{each} embedding.  
This is a natural consequence of the fact that different DR techniques map the same dataset onto different low-dimensional spaces with different sizes. 
As a result, it is possible to achieve any desired ordering of the DR techniques, as demonstrated in~\autoref{fig:teaser} and further details in~\autoref{sec:metrics}.
We study the impact of scale on various stress-based measures~\cite{kruskal1964multidimensional} and propose a safe and reliable variant. 

{
The graph visualization literature also uses stress, and there have been some passing mentions of its scale-sensitivity~\cite{DBLP:journals/tvcg/GansnerHN13,DBLP:journals/cgf/KruigerRMKKT17,DBLP:journals/cgf/WelchK17}. Buja et al.~\cite{buja2008data} were also aware of the scaling problem as early as 2008.  However, the scale-sensitivity property for DR methods does not seem to be widely known as recent implementations do not take scale into account.
}
In particular, our contributions are: 
\begin{compactitem}
    \item Showing analytically and empirically that normalized stress is not bounded to $[0-1]$ and is adversely affected by scale
    \item Reviewing scale-invariant distance-based measures and proposing a scale-invariant stress variant
    \item Empirically showing that  normalized stress on a benchmark dataset can lead to clearly incorrect conclusions
    \item Demonstrating that using a scale-invariant stress measure impacts the results of previous work
\end{compactitem}
These contributions are shown visually in an interactive webpage.\footnote{\href{https://kiransmelser.github.io/normalized-stress-is-not-normalized}{\texttt{kiransmelser\discretionary{}{.}{.}github\discretionary{}{.}{.}io\discretionary{/}{}{/}normalized\discretionary{}{-}{-}stress\discretionary{}{-}{-}is\discretionary{}{-}{-}not\discretionary{}{-}{-}normalized}}}



\section{Background and Related Work}
We start with the existing literature on DR techniques, specifically focusing on MDS and t-SNE. Both of these techniques have been widely studied from a theoretical and mathematical viewpoint~\cite{DBLP:journals/corr/SorzanoVP14,DBLP:journals/ijautcomp/Yin07,DBLP:journals/jmlr/CunninghamG15,DBLP:conf/vluds/EngelHH11}, as well as for their effectiveness in visualizing high-dimensional datasets~\cite{laurens2008dimensionality}. Then, we briefly discuss some quality metrics used to evaluate these techniques and also address a less explored aspect: the impact of scaling the low-dimensional projected data on these metrics.

DR techniques have become popular in the graph layout setting as they can produce more desirable results than traditional force-directed algorithms for some graph structures. 
Kruiger et al.~\cite{DBLP:journals/cgf/KruigerRMKKT17} demonstrated the adaptation of the well-known t-SNE technique for graph layout. Further extensions of these efforts show that t-SNE for graph data also maintains quality with respect to normalized stress and neighborhood preservation~\cite{DBLP:journals/ijdsa/XiaoHH23}.



In some cases, reducing the dimensions of a dataset can improve accuracy for ML tasks such as clustering, as shown 
by Allaoui et al.~\cite{DBLP:conf/icisp/AllaouiKC20}
with Uniform Manifold Approximation and Projection (UMAP). 
It is known that misinterpretations can easily occur with DR plots; for instance, t-SNE can create the illusion of non-existent clusters~\cite{wattenberg2016how}.
For a comprehensive overview of MDS, 
see the work by Hout et al.~\cite{hout2013multidimensional}.

\subsection{Definitions}
We use the following definitions and notations, also found in
Espadoto et al.~\cite{DBLP:journals/tvcg/EspadotoMKHT21},
with an emphasis on visualization. DR is the act of performing a transformation on a collection of objects (observations) found in a $n$-dimensional space (typically, $n \geq 3$) and representing them in a lower $t$-dimensional space (typically, $t = 2$) while preserving some properties of the original data. 

A rectangular matrix $X \in \mathbb{R}^{N \times n}$ represents this collection of $N$ high-dimensional objects, with its rows corresponding to the positions of objects in this space.
The DR transformation results in a rectangular matrix $P \in \mathbb{R}^{N \times t}$ called the projection. 
We can visualize $P$ using a scatter plot or a similar visual idiom.

The $i^{\text{ }th}$ row of matrix $X$, denoted as $x_i$, represents the position of object $i$ in the $n$-dimensional space. Its corresponding position in low-dimensional space is $p_i$. 
{We refer to the pairwise distance between data points $x_i$ and $x_j$ in the high-dimensional space as $\Delta^n(x_i, x_j)$, and we indicate the pairwise distance between the corresponding embedded points $p_i$ and $p_j$ in the low-dimensional space by $\Delta^t(p_i, p_j)$.}
Typically, these are the Euclidean distances in the corresponding dimension.

\begin{table*}[ht]
    \centering
    \renewcommand{\arraystretch}{1.5}
    \caption{Overview of the stress metrics under consideration.}
    \begin{tabular}{| c c c c |}
         \hline 
         Name & Formula & Scale invariant? & References\\ [0.5ex] \hline
         Raw stress & $\sum_{i,j} [\Delta^n(x_i,x_j) - \Delta^t(p_i,p_j)]^2$ & No & \cite{torgerson1952multidimensional}\\ [1ex] \hline 
         Normalized stress & $\frac{\sum_{i,j}[\Delta^n(x_i,x_j) - \Delta^t(p_i,p_j)]^2}{\sum_{i,j}\Delta^n(x_i,x_j)^2}$ & No & 
         \parbox[c]{2cm}{\centering \cite{DBLP:journals/corr/abs-2403-05882,DBLP:journals/bspc/DasanP21,DBLP:journals/corr/abs-2003-09017,DBLP:conf/sibgrapi/MarcilioEG17,DBLP:conf/apvis/AmorimBNSS14,DBLP:conf/ieeevast/AmorimBDJNS12,DBLP:journals/tvcg/PaulovichNML08,DBLP:journals/ivs/NevesFMFP18,DBLP:journals/cgf/PaulovichEPBMN11,DBLP:conf/eurova-ws/MachadoT023,DBLP:conf/igarss/ChenCG06,DBLP:conf/sigmod/FaloutsosL95,DBLP:journals/tvcg/NonatoA19, DBLP:conf/pacificvis/MillerHNHK24,hossain2020multi}}\\ [2ex] \hline 
         Shepard goodness & Spearman correlation $\Delta^n(x_i,x_j), \Delta^t(p_i,p_j)$ & Yes & \cite{DBLP:journals/tvcg/JoiaCCPN11,DBLP:conf/eurova-ws/MachadoT023,DBLP:journals/tkdd/ConnorV24,DBLP:journals/ivs/NevesFMFP18,DBLP:journals/cgf/PaulovichEPBMN11}\\ [0.5ex]      \hline   
         Non-metric stress & $\frac{\sum_{i,j}[\Delta^n(\hat{x}_i,\hat{x}_j) - \Delta^t(p_i,p_j)]^2}{\sum_{i,j}\Delta^t(p_i,p_j)^2}$ & Yes & \cite{DBLP:journals/tkdd/ConnorV24,DBLP:journals/jcc/Agrafiotis03}\\ [2ex] \hline 
         Scale-normalized stress & $\min\limits_{\alpha > 0}\frac{\sum_{i,j}[\Delta^n(x_i,x_j) - \alpha \Delta^t(p_i,p_j)]^2}{\sum_{i,j}\Delta^n(x_i,x_j)^2}$ & Yes & [new]\\ [1.5ex]
         \hline 
         
    \end{tabular}
    \newline
    \label{tab:metrics}
\end{table*}




\subsection{DR Techniques}\label{sec:techniques}
There are many DR techniques that are widely used, and covering them all is outside the scope of this paper. We briefly mention some of the most popular techniques and go into further detail about  MDS and t-SNE, as these are the focus of our study.

  Principal Component Analysis (PCA)~\cite{DBLP:books/sp/Jolliffe86} is among the earliest DR techniques. It is a statistical procedure that uses an orthogonal transformation to convert a set of observations into the so-called principal components. 
  PCA is widely used in exploratory data analysis and for making predictive models.
%
Based on spectral theory, Isomap (ISO)~\cite{doi:10.1126/science.290.5500.2319} is a nonlinear DR method that embeds the dataset into a lower-dimensional space. It is among the first methods in solving the manifold learning problem and finds broad applications in many fields.
  Locally Linear Embedding (LLE)~\cite{doi:10.1126/science.290.5500.2323} is another powerful non-linear method that uses manifold learning techniques for reducing dimensions. The technique uses a locally linear approach as opposed to PCA, which is a global approach. 
  LLE excels in handling problems with spread-out non-linearities, such as in facial image datasets.
  %
  {UMAP~\cite{DBLP:journals/corr/abs-1802-03426}, a recent addition to manifold learning DR techniques, serves as a useful tool for visualization.}
  It shares many similarities with t-SNE (\autoref{sec:tsne}), 
 but is more computationally efficient. 

\subsubsection{MDS}
%
Multi-dimensional Scaling (MDS) refers to a family of DR techniques, each sharing some common characteristics. An input dissimilarity matrix, $D \in \mathbb{R}^{N \times N}$, represents a set of $N$ objects, with the cell $d_{i,j}$ denoting how dissimilar objects $i$ and $j$ are by some metric. The output of any MDS technique is a low-dimensional embedding (typically 2), $P \in \mathbb{R}^{N \times 2}$, which closely preserves these input dissimilarities in some fashion.

There are three main variants of MDS.
Classical scaling~\cite{torgerson1952multidimensional}, otherwise known as Torgerson scaling and Torgerson-Gower scaling, is fundamentally driven by the presumption that the dissimilarities are distances and coordinates can be found that explain them. Because of this assumption, classical scaling admits a closed-form solution, making it efficient and consistent, but suffers from the poor quality of many other linear techniques~\cite{DBLP:journals/tvcg/EspadotoMKHT21}.


{Shepard~\cite{shepard1962analysis} and Kruskal~\cite{kruskal1964multidimensional} define non-metric MDS.}
Non-metric MDS is interested in both finding an embedding of the points in low-dimensional space and a non-parametric relationship between them. At each step of the technique, a monotonic curve of best fit must be found in the drawing’s Shepard diagram, a plot with embedded distances on the $y$\nobreakdash-axis and high-dimensional distances on the $x$\nobreakdash-axis. The smaller the horizontal distances between the curve and points of the diagram are, the better the quality of the embedding is; see~\autoref{fig:MDS-shepard} and~\autoref{fig:random-shepard}. Note that instead of exact distances, this variant only attempts to preserve the ranks between points.
The non-metric stress of Kruskal (defined later in~\autoref{sec:kruskal}) measures the average proportion of how much distance each point-point pair needs to move in the low-dimensional space to satisfy its ranking in the high-dimensional space.

Finally, metric MDS is a collection of techniques which try to directly match high dimensional distances to embedded distances between pairs of points. This includes techniques such as Sammon mapping~\cite{sammon1969nonlinear}, stress majorization~\cite{gansner2005graph}, local MDS~\cite{chen2009local}, and many other techniques.


In short, MDS aims to represent high-dimensional data in a lower-dimensional space while preserving the pairwise distances between data points as much as possible.

\subsubsection{t-SNE}
\label{sec:tsne}
t-SNE~\cite{van2008visualizing} is among the most popular techniques for DR in the field of visualization. Unlike MDS, t-SNE focuses on preserving the local structure of the data. It uses a probabilistic approach to model the similarity between data points in high-dimensional and low-dimensional spaces by paying attention only to nearby points and ignoring far-away points. 

The t-SNE method converts the distances between data points into a Student t-distribution with one degree of freedom. 
{We can represent the similarity between two data points by the conditional probability of the points being neighbors, assuming we select neighbors based on the proportion of their probability density under the distribution~\cite{van2008visualizing}.}



\begin{figure*}[ht]
    \centering
    \includegraphics[width=0.8\linewidth]{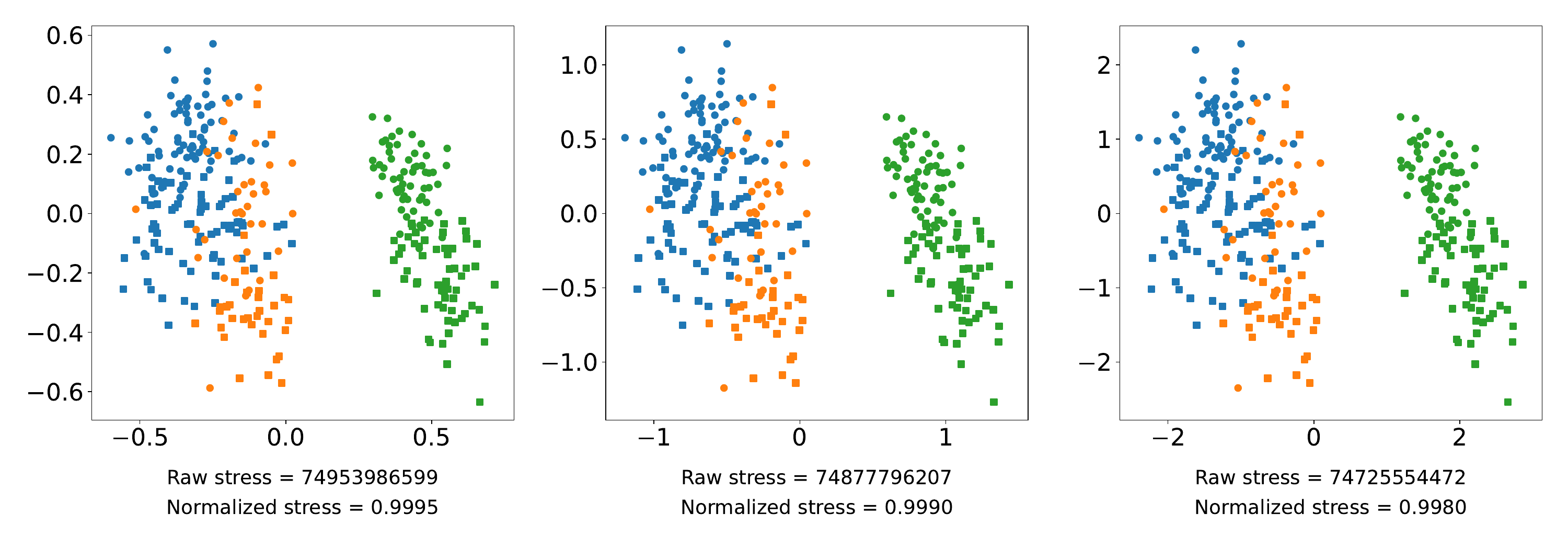}  
    \caption{The $Palmer\text{ }Penguins$ dataset captured by MDS, with species encoded by color and sex encoded by shape. Projections are scaled (by $0.5$, $1$, $2$)  and the value of the metric changes at the different scales.}
    \label{fig:scaled-embeddings}
\end{figure*}

\begin{figure*}[ht]
    \begin{center}
        \includegraphics[width=.8\linewidth]{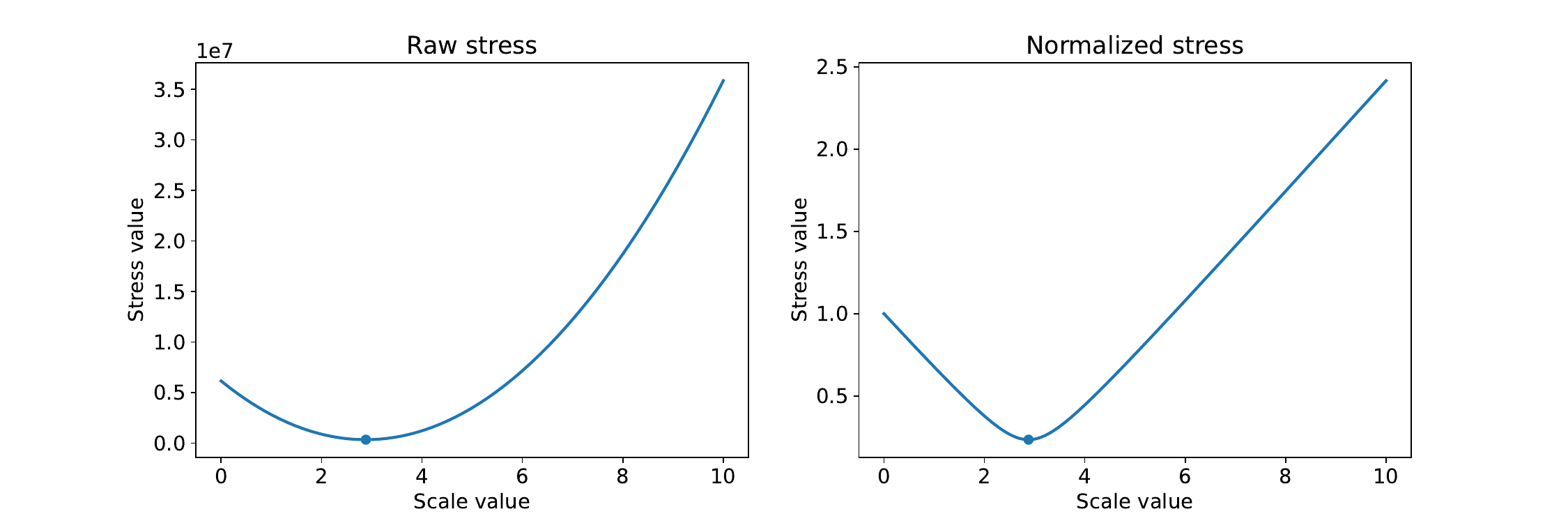}        
    \end{center}
    \vspace{-0.5cm}
    \hspace{0.33\linewidth}(a) \hspace{0.31\linewidth}(b)

    \caption{Plots of the scale-sensitive metrics as one varies the size of the embedding space. Each curve is derived from the $S\text{-}curve$ dataset embedded using MDS. The minimum point has been marked on both curves.}
    \label{fig:norm-stress-curves}
\end{figure*}

\subsection{Quality Metrics and Scale}\label{sec:metrics}
Embedding a sufficiently complex high-dimensional dataset into a lower-dimensional space unavoidably causes distortion, which results in the loss of some information in the dataset~\cite{DBLP:conf/beliv/DasguptaK12}.
The  information lost in the low-dimensional embedding may be distances, neighborhoods, clusters, or other aspects present in the true high-dimensional data. 
Thus, reducing the dimensionality of a dataset may result in the combination or complete removal of features.
Distortion types include stretching, compression, gluing, and/or tearing~\cite{DBLP:journals/ijon/Aupetit07}.

These distortions are difficult to detect visually and decrease the reliability of further analysis conducted on such DR embeddings.  
Then, it is desirable to assess the accuracy of embeddings quantitatively. 
Typically, we accomplish this through the use of quality metrics.

There are many quality metrics used to evaluate the distortion of a low-dimensional embedding, and they each differ in what exactly they compute. 
{We can categorize these metrics as either local measures, global measures, or cluster-level measures~\cite{DBLP:conf/visualization/JeonCJLHKJS23}.}

Local measures gauge the preservation of the neighborhood structure of the high-dimensional dataset in the low-dimensional embedding.
This includes measures such as Trustworthiness \& Continuity~\cite{DBLP:journals/nn/VennaK06}, Mean Relative Rank Errors (MRRE)~\cite{DBLP:journals/ijon/LeeV09}, and Neighborhood Hit~\cite{DBLP:journals/tvcg/PaulovichNML08}. 

Cluster-level measures assess the maintenance of the cluster structures inherent in the high-dimensional data in the low-dimensional embedding.
This includes  Steadiness \& Cohesiveness~\cite{DBLP:journals/tvcg/JeonKJKS22} and Distance Consistency~\cite{DBLP:journals/cgf/SipsNLH09}. 

Global measures evaluate the degree to which pairwise distances between objects remain consistent between the high- and low-dimensional data. 
All stress-based measures in~\autoref{tab:metrics} fall under this classification and are described in~\autoref{sec:stress}.  

The low-dimensional embeddings of a dataset by different DR techniques often lead to the original high-dimensional data being projected into distinct low-dimensional spaces. The size or bounding box of these spaces may vary greatly depending on the technique employed. For instance, a technique designed for visualization might prefer to output the units as pixel coordinates, which will be a large region compared to another technique designed for ML, which outputs coordinates between 0 and 1. 

This becomes a problem when an embedding evaluation metric is not scale-invariant, meaning that the value of the metric changes when the size of the output changes. Just as translation and rotation, uniformly scaling up or down, does not meaningfully change the relationships present in an embedding; see~\autoref{fig:scaled-embeddings}. 

For this reason, when computing the score of a quality metric that is susceptible to scale, one must be careful when comparing the quality of two different embeddings produced by different DR techniques.
If the scale of one low-dimensional embedding is significantly smaller or larger than the scale of another low-dimensional embedding, this difference may lead to unreasonably high values of normalized stress, indicating poor preservation of the original high-dimensional data structure. 

{The three stress-scale curves plotted in~\autoref{fig:teaser} demonstrate this.}
Computing normalized stress for a given embedding without accounting for scale corresponds to selecting an arbitrary point along its curve (as scale is the $x$\nobreakdash-coordinate). 
{When we consider a second embedding, we generate a second curve and select another arbitrary scale.}
Notably, the second scale may be \textit{different} from the first scale. In terms of the curves in~\autoref{fig:teaser}, this means we may have selected $x=2$ for $MDS$ but $x=0.5$ for \textit{t-SNE} (implying \textit{t-SNE} has captured distances better). Hence, without accounting for scale, one can establish any desired ordering between different DR techniques, making it crucial to take scale into account when comparing the performance of DR techniques to ensure that the quality metric used accurately reflects the performance of the techniques.

\section{Stress}\label{sec:stress}
Stress is one of
the most commonly utilized metrics 
to gauge the quality of low-dimensional projections using DR techniques~\cite{DBLP:journals/tvcg/EspadotoMKHT21,DBLP:journals/tvcg/NonatoA19,DBLP:conf/sigmod/FaloutsosL95,DBLP:journals/tvcg/PaulovichNML08}.
{Shepard first introduced stress in the field of psychometrics in the 1960s, and Kruskal later expanded on it~\cite{kruskal1964multidimensional,kruskal1964nonmetric,shepard1962analysis}. While Kruskal's definition is in fact scale invariant, the normalized stress metric that has come to be widely used is not.}
The goal of these papers was to develop a DR technique that 
maintained the pairwise distances between data points as well as possible. 

{Researchers have proposed different stress variants as objective functions to achieve this type of embedding. In general, a stress measure, denoted as a real-valued function $M: (\mathbb{R}^{N \times n}, \mathbb{R}^{N \times t}) \rightarrow \mathbb{R}$, 
maps the pair of a high-dimensional matrix $X$ and its embedding $P$ to a real number.}

\subsection{Scale-sensitive stress measures}
As discussed previously, the most popular stress measures are scale-sensitive. A scale-sensitive measure is a measure in which the value changes as the size of the projection changes. 
In other words, for a metric $M(X,\alpha P)$, the result is a non-constant function of $\alpha$. 

\begin{figure*}[ht]
    \centering
    \includegraphics[width=0.6\linewidth]{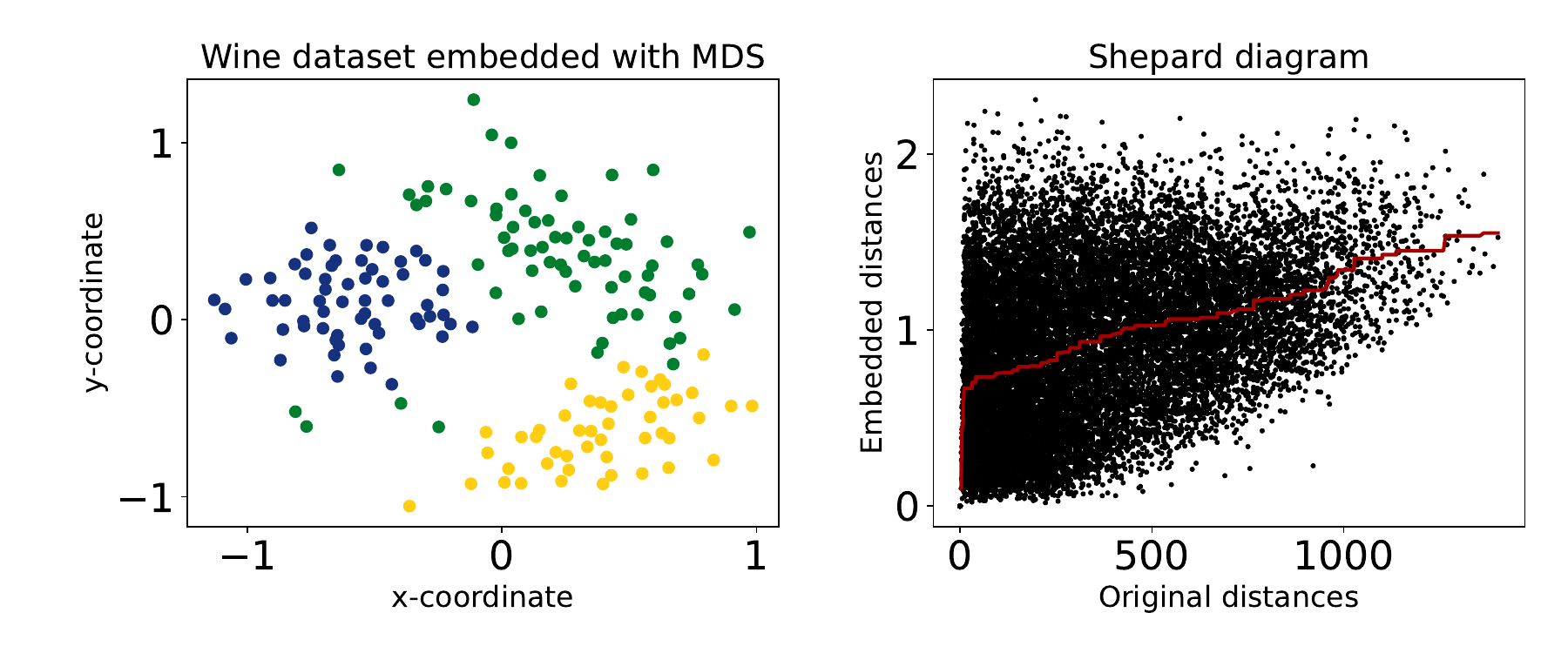}
    \caption{Embedding and Shepard diagram with monotonic fitted line for the $Wine$ dataset captured by MDS.}
    \label{fig:MDS-shepard}
\end{figure*}

\begin{figure*}[ht]
    \centering
    \includegraphics[width=0.6\linewidth]{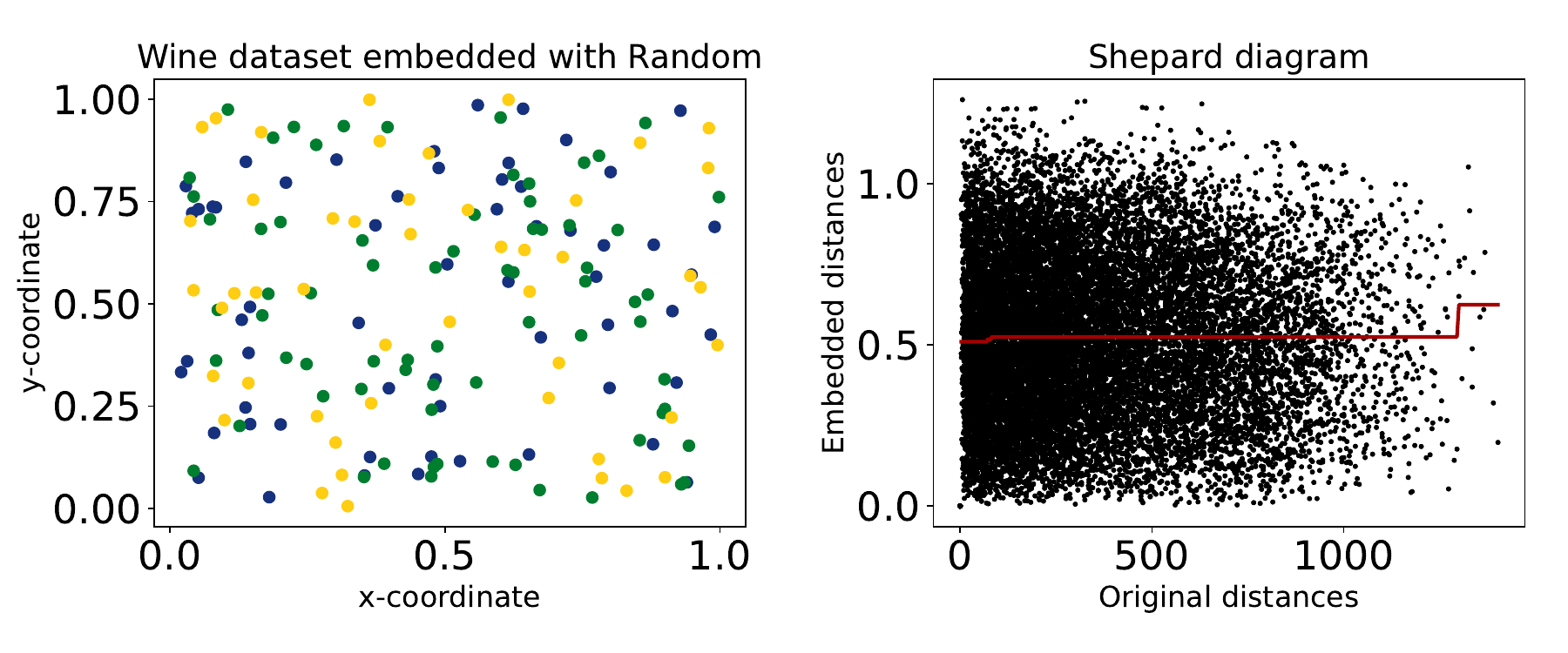}
    \caption{Embedding and Shepard diagram with monotonic fitted line for the $Wine$ dataset captured by RND.}
    \label{fig:random-shepard}
\end{figure*}

\subsubsection{Raw Stress (RS)}
The first measure that arose is the measure of goodness of fit, referred to as raw stress (RS), which is the difference between distances in the high-dimensional space and distances in the low-dimensional space. This first appeared in Torgerson's calssical MDS paper~\cite{torgerson1952multidimensional}. RS is formally defined as follows,
\begin{equation}
    \label{eq:raw-stress}
    \text{RS(X,P)}=\sum_{i,j}[\Delta^n (x_i, x_j)-\Delta^t (p_i, p_j)]^2.
\end{equation}
However, although raw stress is invariant under distortions such as rotations, translations, and reflections, it is variant under stretching and compression distortions~\cite{kruskal1964multidimensional}. We can show this analytically: 
\[\sum_{i,j}[\Delta^n (x_i, x_j)- \Delta^t (\alpha p_i, \alpha p_j)]^2 = \sum_{i,j}[\Delta^n (x_i, x_j)- \alpha\Delta^t (p_i, p_j)]^2\]
\[\sum_{i,j}\Delta^n (x_i, x_j)^2 + \alpha\sum_{i,j}\Delta^n (x_i, x_j)\Delta^t (p_i, p_j) + \alpha^2\sum_{i,j}\Delta^t (p_i, p_j)^2\text{.}\]
We see that we can rewrite the raw stress as a quadratic function of $\alpha$.
This behavior is obvious in
{~\autoref{fig:norm-stress-curves}(a).}

\subsubsection{Normalized Stress (NS)}
{
RS can produce very large stress values even for reasonably good embeddings, making it not ideal to use as an evaluation metric. For this reason, the normalized stress (NS) measure has become more popular to use as the range of possible values tends to be much smaller. Likely the first instance of normalized stress is its use in Sammon mapping~\cite{sammon1969nonlinear}, a type of metric MDS. In its original context, normalized stress was not intended to be used as an evaluation metric, but as an optimization function.
We refer to the square root of the sum of squared differences between the original and reduced distances, divided by the sum of squared original distances, as normalized stress (NS)~\cite{DBLP:journals/tvcg/EspadotoMKHT21}.
}
\begin{equation}
    \label{eq:norm-stress}
    \text{NS(X,P)}=\sqrt{\frac{\sum_{i,j}[\Delta^n (x_i, x_j)-\Delta^t (p_i, p_j)]^2}{\sum_{i,j}\Delta^n (x_i, x_j)^2}}
\end{equation}
Taking the square root reduces the impact of large residuals and emphasizes smaller ones (but does not change any ordering). This measure ensures that the stress value is more sensitive to smaller discrepancies between the original and reduced distances, which can be crucial when measuring the preservation of the structure of the data during DR. 

Just as raw stress, however, normalized stress is scale-sensitive, as illustrated in~\autoref{fig:norm-stress-curves}(b).
Note that although one might assume that normalized stress has values in the range $[0,1]$ (e.g., as Espadoto et al.~\cite{DBLP:journals/tvcg/EspadotoMKHT21}), this is indeed not true. As $\Delta^t(p_i,p_j)$ is unbounded, the range of normalized stress is also unbounded. 

Scaling the term $\Delta^t(p_i,p_j)$ by $0$ results in a normalized stress value of $1$, because $\Delta^t(p_i,p_j)=0$.
Thus, the normalized stress curve of any DR technique (with respect to scale) initially starts at $y=1$, decreases until reaching a minimum value, and then increases in unbounded fashion. 
The quadratic nature of the normalized stress function, defined in~\autoref{eq:norm-stress}, causes this behavior.
Hence, the normalized stress curves of pairs of DR techniques have a common intersection at $x=0$, as illustrated in~\autoref{fig:teaser}, and possibly one more positive intersection further to the right.


\subsection{Scale-invariant Stress Measures}
While the most popular stress measures are scale-sensitive, there are indeed some scale-invariant ones, i.e., they are constant with respect to $\alpha$ for a measure $M(X,\alpha P)$.

\subsubsection{Shepard Goodness Score (SGS)}
One variant of stress introduced in 1962 is now known as the Shepard goodness score (SGS)~\cite{shepard1962analysis}. It has two  components: the Shepard diagram and the Spearman rank correlation.

The Shepard diagram is a scatterplot of all $\binom{N}{2}$ pairwise distances between points in the high-dimensional space versus the corresponding pairwise distances in the low-dimensional space.
In other words, the Shepard diagram is a collection of two-dimensional points. 
\[c_{i,j} = (\Delta^n(x_i,x_j), \Delta^t(p_i,p_j)) \; \forall 1 \leq i < j \leq N\]

This can be determined by visual assessment: as the scatter plot approaches a positive diagonal line, the preservation of distances between the two spaces improves~\cite{DBLP:journals/tvcg/EspadotoMKHT21}.~\autoref{fig:MDS-shepard} shows an example of a projection with a good Shepard diagram, and~\autoref{fig:random-shepard} shows a projection with a bad Shepard diagram.

The Shepard goodness score is then the Spearman rank correlation of the $x$ and $y$ components of the diagram, i.e., the smallest distance in high-dimensional space should still be the smallest in the projection, and the $k$ smallest distance should still be the $k$ smallest. 
{The rank correlation, being a unit-less measure based on the variance of the ranks, remains invariant to scale.}
Stretching or shrinking the projection uniformly will not change the ranking of distances. 

\subsubsection{Non-metric stress (NMS)}
\label{sec:kruskal}
Another variant of stress proposed in 1964 is non-metric stress (NMS), sometimes referred to as Kruskal stress~\cite{kruskal1964multidimensional}. Non-metric stress does not measure the preservation of the pairwise distances between points in high-dimensional and low-dimensional spaces. Instead, it measures how well the ranking (or ordering) of the distances is maintained between the spaces. Formally, non-metric stress is defined as follows:
\begin{equation}
    \label{eq:non-metric-stress}
    \text{NMS(X,P)}=\frac{\sum_{i,j}[\Delta^n(\hat{x}_i,\hat{x}_j) - \Delta^t(p_i,p_j)]^2}{\sum_{i,j}\Delta^t(p_i,p_j)^2},
\end{equation}
where the matrix $\hat{X}$ is computed by first using the Shepard diagram, as previously described, to obtain pairwise distances between all points, and then a monotonically increasing line of best fit (isotonic regression) on $\binom{n}{2}$ degrees of freedom is also computed. Finally, $\hat{X}$ is populated with the differences from the pairwise distances to the fitted line in only the $y$\nobreakdash-coordinate. 
Both~\autoref{fig:MDS-shepard} and~\autoref{fig:random-shepard} demonstrate this monotonic curve.

It is not immediately obvious why this metric is scale-invariant. 
A closer examination shows the impact of the new term $\hat{X}$, defined by the vertical distance to the fitted line. 
Scaling a projection by $\alpha$ affects only the $y$\nobreakdash-axis of the Shepard diagram.
Thus, when scaled: 
\[NMS(X,\alpha P) = \frac{\sum_{i,j}[\alpha \Delta^n(\hat{x}_i,\hat{x}_j) - \alpha \Delta^t(p_i,p_j)]^2}{\sum_{i,j}(\alpha \Delta^t(p_i,p_j))^2}\]
\[ = \frac{\alpha^2 \sum_{i,j}[\Delta^n(\hat{x}_i,\hat{x}_j) - \Delta^t(p_i,p_j)]^2}{\alpha^2 \sum_{i,j}\Delta^t(p_i,p_j)^2}\]
The $\alpha$ factors cancel, and the metric is scale invariant.






\subsubsection{Scale-normalized Stress (SNS)}\label{sec:SNS}
As we have already noted theoretically and empirically, both the raw stress and normalized stress metrics are scale-sensitive. We also showed analytically that they are, in fact, quadratic functions of the scaling factor $\alpha$. This lends itself to the nice property that there is a unique minimum (since the leading term must be positive).  
As a result, these functions are parabolic in the scaling factor $\alpha$; see~\autoref{fig:teaser} and~\autoref{fig:norm-stress-curves}.
%
With this in mind, we propose the {\em scale-normalized stress (SNS)} as a `quick fix' for the normalized scale metric:
\begin{equation}
    \label{eq:scale-norm-stress}
    SNS(X,P) = \min\limits_{\alpha > 0} \; NS(X,\alpha P)
\end{equation}



This produces the ``true” normalized stress score, as it gives us scale invariance and is straightforward to explain. 
%
We note that the scale-variance of stress-based metrics has been observed in the graph visualization literature, and scale-normalized stress values were computed~\cite{DBLP:journals/tvcg/GansnerHN13,DBLP:journals/cgf/KruigerRMKKT17}. 
We  next show how to compute the minimum directly and efficiently. 

To find the value of $\alpha$ that minimizes the normalized stress function (\autoref{eq:norm-stress}), we take the derivative of the numerator with respect to $\alpha$ and set it equal to zero. This is given by:
\begin{gather*}
    \frac{d}{d\alpha} \sum_{i,j}[\Delta^n (x_i, x_j)-\alpha\Delta^t (p_i, p_j)]^2 = 0.
\end{gather*}
Solving this equation gives us the optimal value of $\alpha$:
\begin{gather*}
    \alpha = \frac{\sum_{i,j}[\Delta^n (x_i, x_j)*\Delta^t (p_i, p_j)]}{\sum_{i,j}\Delta^t (p_i, p_j)^2}.
\end{gather*}

The denominator in the $\alpha$ equation normalizes the distances, so that the scale of the distances in the high-dimensional space does not affect the value of $\alpha$. The numerator is essentially a covariance between the high-dimensional distances and the low-dimensional distances, ensuring that $\alpha$ captures the relationship between the two sets of distances. 
{We can interpret the equation for $\alpha$ as a correlation coefficient between the high-dimensional and low-dimensional distances.}
An implementation that calculates the aforementioned scale-normalized stress metric is available in our code \hyperlink{https://github.com/KiranSmelser/dim-reduction-metrics-and-scale}{repository}.


\section{Numerical Experiments}

In this section, we demonstrate, beyond the few examples shown up to this point, the frequency at which stress metrics behave in an expected manner and agree with one another when comparing the performance of DR techniques. The goal is to identify the stress metrics that most often agree as well as consistently rank embeddings from high-quality DR techniques better than embeddings from poor-quality DR techniques. 
In particular, we have three questions we would like to answer through empirical experiments: 
\begin{compactenum}
    \item[\textbf{Q1}]: Which metrics most often agree with the expected order of techniques? 
    \item[\textbf{Q2}]: Which of these metrics most often agree with each other?
    \item[\textbf{Q3}]: Have scale-sensitive metrics effected the results of previous DR evaluations? 
\end{compactenum}

{We designed \hyperref[sec:experiment-A]{Experiment A} to answer \textbf{Q1} and \textbf{Q2}, and \hyperref[sec:experiment-B]{Experiment B} to address \textbf{Q3}. Below, we describe the datasets, DR techniques, and quality metrics that we used in the experimentats.}

The code to conduct the following experiments was written in Python and made use of the sklearn, umap, scipy, and ZADU libraries and is available.\footnote{\href{https://github.com/KiranSmelser/stress-metrics-and-scale}{\texttt{kiransmelser\discretionary{}{.}{.}github\discretionary{}{.}{.}io\discretionary{/}{}{/}stress\discretionary{}{-}{-}metrics\discretionary{}{-}{-}and\discretionary{}{-}{-}scale}}}
An accompanying webpage to help  explore the behavior of normalized stress is also available.\footnote{\href{https://kiransmelser.github.io/normalized-stress-is-not-normalized}{\texttt{kiransmelser\discretionary{}{.}{.}github\discretionary{}{.}{.}io\discretionary{/}{}{/}normalized\discretionary{}{-}{-}stress\discretionary{}{-}{-}is\discretionary{}{-}{-}not\discretionary{}{-}{-}normalized}}}

\subsection{Datasets}

\begin{table}[ht]
\centering
\caption{\label{tab:datasets}Dataset statistics.}
\small 
\begin{tabular}{|c c c c|}
\hline
Dataset & Type & Size ($N$) & Dimensionality ($n$) \\
\hline
iris & tabular & 149 & 4 \\
\hline
wine & tabular & 178 & 13 \\
\hline
swiss roll & synthetic & 1500 & 3 \\
\hline
penguins & tabular & 342 & 4 \\
\hline
auto-mpg & multivariate & 391 & 5 \\
\hline
s-curve & synthetic & 1500 & 3 \\
\hline
\hline
bank & multivariate & 45211 & 16 \\
\hline
cifar10 & image & 60000 & 1024 \\
\hline
coil20 & image & 1440 & 16384 \\
\hline
fashion\_mnist & image & 3000 & 784 \\
\hline
fmd & image & 997 & 1536 \\
\hline
har & multivariate & 735 & 561 \\
\hline
hatespeech & text & 3222 & 100 \\
\hline
hiva & tabular & 3076 & 1617 \\
\hline
imdb & text & 3250 & 700 \\
\hline
orl & image & 400 & 396 \\
\hline
secom & multivariate & 1567 & 590 \\
\hline
seismic & multivariate & 646 & 24 \\
\hline
sentiment & text & 2748 & 200 \\
\hline
sms & text & 836 & 500 \\
\hline
spambase & text & 4601 & 57 \\
\hline
svhn & image & 733 & 1024 \\
\hline
\end{tabular}
\vspace{0.15cm}
\vspace{-0.5cm}

\end{table}

We selected the datasets for our experiments to cover a wide range of characteristics and complexities.
They include the well-known \textit{Iris}~\cite{Unwin2021TheID}, \textit{Wine}~\cite{DBLP:journals/pr/AeberhardCV94}, \textit{Swiss Roll}~\cite{DBLP:books/daglib/0029547}, \textit{Palmer Penguins}~\cite{10.1371/journal.pone.0090081}, \textit{Auto MPG}~\cite{DBLP:data/10/Quinlan93}, and \textit{S-Curve}~\cite{DBLP:journals/jmlr/PedregosaVGMTGBPWDVPCBPD11} datasets. In addition to these, we have also included all 16 datasets from the recent survey of DR techniques
by Espadoto et al.~\cite{DBLP:journals/tvcg/EspadotoMKHT21}. 
\autoref{tab:datasets} details some statistics about these datasets.
%
The comprehensive corpus of datasets we use allows us to evaluate the behavior of stress under various conditions.

\subsection{Techniques}
We experimented with several different DR techniques on each of the datasets described above and identified three—MDS, t-SNE, and the random projection technique (RND)—that most effectively highlighted the differences in scale. (The results of all the DR techniques can be found in our code \hyperlink{https://github.com/KiranSmelser/dim-reduction-metrics-and-scale}{repository}.) 
\begin{compactenum}
  \item MDS directly optimizes the normalized stress function, so the stress score of the embedding generated should be low.
  \item t-SNE emphasizes maintaining the local structure of the data. Therefore, it should produce higher stress scores than MDS.
  \item RND assigns every point $(x,y)$ coordinates at random in the 2D unit square. The resulting embedding has no structure and consequently should have the highest stress score.
\end{compactenum}
In terms of the stress-based quality of the embeddings, we anticipate that the “correct order” of these three techniques should be as follows: MDS, t-SNE, and RND (from better to worse).
\subsection{Metrics}
The quality metrics used to assess the chosen DR techniques throughout the following experiments include the raw stress score, normalized stress score, scale-normalized stress score, Shepard goodness score, and non-metric stress score; see~\autoref{tab:metrics}.

We use the implementations of normalized stress  and  Shepard Goodness from the ZADU library~\cite{DBLP:conf/visualization/JeonCJLHKJS23}. 
{We implemented all remaining stress measures, based on the definitions provided in~\autoref{sec:stress}, the code for which is made available.}

\section{Empirical Experiments}

\subsection{Experiment A: Ground Truth Experiment}\label{sec:experiment-A}
Each of the techniques (MDS, t-SNE, and RND) embeds each dataset from a random initialization $10$ times, resulting in a total of $10 \times 3 \times 22 = 660$ different embeddings. For each projection, we compute each of the five metrics described in
~\autoref{sec:stress}.

An {\em order} is an ordered arrangement of the three techniques, such that the first technique has the lowest metric score, the second technique has the next lowest, and the third technique has the highest score (where lower is better).  


We can address \textbf{Q1} by analyzing the order of techniques for each metric with respect to the quality of an embedding. For a given collection of projections of the same dataset, we obtain an order given by a metric, i.e., the best, middle, and worst projections. We count the number of times these orders occur across our benchmark, paying  attention to the expected order of MDS, t-SNE, and RND.

In order to answer \textbf{Q2}, we assess the agreement between the rankings of DR techniques produced by the stress metrics. This is done by calculating Spearman's rank-order correlation for each pair of quality metrics.
This approach contrasts with our strategy to address \textbf{Q1}: instead of 
documenting whether a metric yields the expected order (represented by $0$ or $1$), Spearman’s rank-order correlation takes into account the degree of similarity or dissimilarity between the produced orderings.

A clear pattern emerges across the large variety of datasets. 
The results for \textbf{Q1} are summarized in~\autoref{tab:results-table}. We can easily see that the scale-sensitive metrics of raw stress and normalized stress fail to capture this expected ordering, only obtaining it less than 1\% of the time. While they never rank a RND projection lower than an MDS projection, they often ($\approx$80\%) do rank t-SNE lower than a RND projection. This is unintuitive, as we expect that while t-SNE does not preserve distances better than MDS, it most certainly should do so more than a RND projection. 
The raw stress and normalized stress metrics also yield identical results in~\autoref{tab:results-table} and~\autoref{tab:results-table10x}, which is mathematically consistent as both metrics are derived from the same stress function, with the normalized stress being a scaled version of the raw stress.

When we increase the size of the projections by a factor of 10, the results for raw and normalized stress change dramatically; see~\autoref{tab:results-table10x}. The most common order ($\approx$80\%) becomes RND$<$MDS$<$t-SNE, with RND, the completely uncorrelated projection, achieving the lowest stress. 

On the other hand, the scale-invariant metrics do a much better job at capturing the expected ordering of projections, with scale-normalized and Shepard goodness scores both achieving greater than 90\% accuracy. This holds true regardless of the size of the projection; the values are the same in~\autoref{tab:results-table} and~\autoref{tab:results-table10x}. 

\begin{table}[ht]
    \centering
    \caption{Percentage of each possible ordering of the three projections (rows) given by each metric (columns). The first row (MDS, t-SNE, RND) is the expected distance preservation ordering. Metrics with high values on this row most often capture this ordering. }
    \begin{tabularx}{\linewidth}{|p{2.65cm}| X X X X X|}
    \hline 
     & RS & NS & SGS & NMS & SNS \\ \hline 
    MDS$<$t-SNE$<$RND & \cellcolor[HTML]{F7FCFD} 4.2\% & \cellcolor[HTML]{F7FCFD} 4.2\% & \cellcolor[HTML]{5AB5D9} 91.7\% & \cellcolor[HTML]{6BBCDD} 82.1\% & \cellcolor[HTML]{5CB5DA} 90.8\%\\ \hline
    MDS$<$RND$<$t-SNE & \cellcolor[HTML]{69BCDD} 83.3\% & \cellcolor[HTML]{69BCDD} 83.3\% & \cellcolor[HTML]{FFFFFF} 0.0\% & \cellcolor[HTML]{DFF1F8} 17.9\% & \cellcolor[HTML]{F0F8FC} 8.3\%\\ \hline
    t-SNE$<$MDS$<$RND & \cellcolor[HTML]{E8F5FA} 12.5\% & \cellcolor[HTML]{E8F5FA} 12.5\% & \cellcolor[HTML]{F0F8FC} 8.3\% & \cellcolor[HTML]{FFFFFF} 0.0\% & \cellcolor[HTML]{FEFEFF} 0.8\%\\ \hline
    t-SNE$<$RND$<$MDS & \cellcolor[HTML]{FFFFFF} 0.0\% & \cellcolor[HTML]{FFFFFF} 0.0\% & \cellcolor[HTML]{FFFFFF} 0.0\% & \cellcolor[HTML]{FFFFFF} 0.0\% & \cellcolor[HTML]{FFFFFF} 0.0\%\\ \hline
    RND$<$MDS$<$t-SNE & \cellcolor[HTML]{FFFFFF} 0.0\% & \cellcolor[HTML]{FFFFFF} 0.0\% & \cellcolor[HTML]{FFFFFF} 0.0\% & \cellcolor[HTML]{FFFFFF} 0.0\% & \cellcolor[HTML]{FFFFFF} 0.0\%\\ \hline
    RND$<$t-SNE$<$MDS & \cellcolor[HTML]{FFFFFF} 0.0\% & \cellcolor[HTML]{FFFFFF} 0.0\% & \cellcolor[HTML]{FFFFFF} 0.0\% & \cellcolor[HTML]{FFFFFF} 0.0\% & \cellcolor[HTML]{FFFFFF} 0.0\%\\ \hline
    \end{tabularx}
    \newline
    \label{tab:results-table}
\end{table}

\begin{table}[ht]
    \centering
    \caption{The same experiment from~\autoref{tab:results-table}, with projections scaled up by 10. Note how, by just changing the size of the projection, raw and normalized stress  consistently rank the RND technique better. }
    \begin{tabularx}{\linewidth}{|p{2.65cm}| X X X X X|}
    \hline 
     & \small{RSx10} & \small{NSx10} & \small{SGSx10} & \small{NMSx10} & \small{SNSx10} \\ \hline
    MDS$<$t-SNE$<$RND & \cellcolor[HTML]{FFFFFF} 0.0\% & \cellcolor[HTML]{FFFFFF} 0.0\% & \cellcolor[HTML]{5AB5D9} 91.7\% & \cellcolor[HTML]{6BBCDD} 82.1\% & \cellcolor[HTML]{5CB5DA} 90.8\%\\ \hline
    MDS$<$RND$<$t-SNE & \cellcolor[HTML]{F0F8FC} 8.3\% & \cellcolor[HTML]{F0F8FC} 8.3\% & \cellcolor[HTML]{FFFFFF} 0.0\% & \cellcolor[HTML]{DFF1F8} 17.9\% & \cellcolor[HTML]{F0F8FC} 8.3\%\\ \hline
    t-SNE$<$MDS$<$RND & \cellcolor[HTML]{E8F5FA} 12.5\% & \cellcolor[HTML]{E8F5FA} 12.5\% & \cellcolor[HTML]{F0F8FC} 8.3\% & \cellcolor[HTML]{FFFFFF} 0.0\% & \cellcolor[HTML]{FEFEFF} 0.8\%\\ \hline
    t-SNE$<$RND$<$MDS & \cellcolor[HTML]{FFFFFF} 0.0\% & \cellcolor[HTML]{FFFFFF} 0.0\% & \cellcolor[HTML]{FFFFFF} 0.0\% & \cellcolor[HTML]{FFFFFF} 0.0\% & \cellcolor[HTML]{FFFFFF} 0.0\%\\ \hline
    RND$<$MDS$<$t-SNE & \cellcolor[HTML]{70BFDF} 79.2\% & \cellcolor[HTML]{70BFDF} 79.2\% & \cellcolor[HTML]{FFFFFF} 0.0\% & \cellcolor[HTML]{FFFFFF} 0.0\% & \cellcolor[HTML]{FFFFFF} 0.0\%\\ \hline
    RND$<$t-SNE$<$MDS & \cellcolor[HTML]{FFFFFF} 0.0\% & \cellcolor[HTML]{FFFFFF} 0.0\% & \cellcolor[HTML]{FFFFFF} 0.0\% & \cellcolor[HTML]{FFFFFF} 0.0\% & \cellcolor[HTML]{FFFFFF} 0.0\%\\ \hline
    \end{tabularx}
    \newline
    \label{tab:results-table10x}
\end{table}

In the context of \textbf{Q2},~\autoref{tab:corrs-table} presents the correlation coefficients.
As expected, the scale-sensitive measures of raw stress and normalized stress will always give the same order, so their correlation is 1. However, they have a weak correlation to the scale-invariant metrics, hovering around 0.5. The scale-invariant metrics have strong correlations between themselves.  

\begin{table}[ht]
    \centering
    \caption{Spearman correlations between stress metrics when considering the ordering of the RND, t-SNE, and MDS embedding techniques. These values are with respect to the orders found in~\autoref{tab:results-table}.}
    \begin{tabularx}{\linewidth}{|p{0.5cm}| X X X X X|}
    \hline 
     & RS & NS & SGS & NMS & SNS\\ \hline
    RS & \cellcolor[HTML]{4BAED6} 1.0 & \cellcolor[HTML]{4BAED6} 1.0 & \cellcolor[HTML]{A4D6EA} 0.507 & \cellcolor[HTML]{A5D6EA} 0.501 & \cellcolor[HTML]{A0D4E9} 0.526\\ \hline
    NS & -- & \cellcolor[HTML]{4BAED6} 1.0 & \cellcolor[HTML]{A4D6EA} 0.507 & \cellcolor[HTML]{A5D6EA} 0.501 & \cellcolor[HTML]{A0D4E9} 0.526\\ \hline
    SGS & -- & -- & \cellcolor[HTML]{4BAED6} 1.0 & \cellcolor[HTML]{66BADC} 0.849 & \cellcolor[HTML]{59B4D9} 0.922\\ \hline
    NMS & -- & -- & -- & \cellcolor[HTML]{4BAED6} 1.0 & \cellcolor[HTML]{55B2D8} 0.947\\ \hline
    SNS & -- & -- & -- & -- & \cellcolor[HTML]{4BAED6} 1.0\\ \hline
    \end{tabularx}
    \newline
    \label{tab:corrs-table}
\end{table}

\subsection{Experiment B: Validation Experiment}\label{sec:experiment-B}
It is also important to demonstrate how reevaluating prior studies using our proposed scale-normalized stress metric rather than normalized stress affects results. Here, we aim to address \textbf{Q3} by revisiting the data from the quantitative survey of DR methods by Espadoto et al.~\cite{DBLP:journals/tvcg/EspadotoMKHT21}. Specifically, we use 5 embeddings (MDS, t-SNE, LLE, ISO, and UMAP) computed for each of the 17 datasets in their corpus and available on the website accompanying their paper. Note that each of these embeddings was finely tuned with hyper-parameters to achieve good results, and the embeddings are at different scales.
We then compute two metrics for each of these embeddings: the normalized stress used in the survey evaluation and the scale-normalized stress proposed in~\autoref{sec:SNS}.

~\autoref{tab:rerun-table-NS} and~\autoref{tab:rerun-table-SNS} present the rankings that normalized stress versus scale-normalized stress produced for the techniques and datasets in Espadoto et al.~\cite{DBLP:journals/tvcg/EspadotoMKHT21}.
The first thing that stands out is that MDS is consistently ranked the best regardless of metric,  which is consistent with the fact that MDS directly optimizes normalized stress. 
However, the order of the other four techniques changes considerably. For instance, on the \textit{fmd} dataset, none of the other four techniques are in the same position, from normalized stress to scale-normalized stress. 

Normalized stress  places ISO as the second best, approximately 41\% of the time, and LLE as the second best (around 35\% of the time). Meanwhile, the scale-normalized stress metric ranks t-SNE as the second best (approximately 70\% of the time).

When considering the techniques that both metrics rank as the least effective, normalized stress ranks t-SNE last around 70\% of the time, while scale-normalized stress ranks LLE last approximately 70\% of the time.

This 
begins to answer \textbf{Q3} and
calls into question the validity of results based on normalized stress as an evaluation metric.

\begin{table}[ht]
    \centering
    \caption{The order given by normalized stress on embeddings of datasets (rows). Each embedding (MDS, t-SNE, UMAP, LLE, and ISO) is according to metric value, with lower ranks having lower stress.}
    \small
    \begin{tabularx}{\linewidth}{|p{1.6cm}| X X X X p{.9cm}|}
    \multicolumn{6}{c}{\textbf{NS Ranking}}\\
    \hline 
     & MDS & ISO & LLE & UMAP & t-SNE\\ \hline
    bank & \cellcolor[HTML]{FFFFFF} 1 & \cellcolor[HTML]{D2EBF5} 2 & \cellcolor[HTML]{A5D6EA} 3 & \cellcolor[HTML]{78C2E0} 4 & \cellcolor[HTML]{4BAED6} 5\\ \hline
    cifar10 & \cellcolor[HTML]{FFFFFF} 1 & \cellcolor[HTML]{D2EBF5} 2 & \cellcolor[HTML]{78C2E0} 4 & \cellcolor[HTML]{A5D6EA} 3 & \cellcolor[HTML]{4BAED6} 5\\ \hline
    cnae9 & \cellcolor[HTML]{FFFFFF} 1 & \cellcolor[HTML]{D2EBF5} 2 & \cellcolor[HTML]{A5D6EA} 3 & \cellcolor[HTML]{78C2E0} 4 & \cellcolor[HTML]{4BAED6} 5\\ \hline
    coil20 & \cellcolor[HTML]{FFFFFF} 1 & \cellcolor[HTML]{78C2E0} 4 & \cellcolor[HTML]{D2EBF5} 2 & \cellcolor[HTML]{A5D6EA} 3 & \cellcolor[HTML]{4BAED6} 5\\ \hline
    epileptic & \cellcolor[HTML]{FFFFFF} 1 & \cellcolor[HTML]{D2EBF5} 2 & \cellcolor[HTML]{A5D6EA} 3 & \cellcolor[HTML]{4BAED6} 5 & \cellcolor[HTML]{78C2E0} 4\\ \hline
    fashion\_mnist & \cellcolor[HTML]{FFFFFF} 1 & \cellcolor[HTML]{78C2E0} 4 & \cellcolor[HTML]{A5D6EA} 3 & \cellcolor[HTML]{D2EBF5} 2 & \cellcolor[HTML]{4BAED6} 5\\ \hline
    fmd & \cellcolor[HTML]{FFFFFF} 1 & \cellcolor[HTML]{D2EBF5} 2 & \cellcolor[HTML]{78C2E0} 4 & \cellcolor[HTML]{4BAED6} 5 & \cellcolor[HTML]{A5D6EA} 3\\ \hline
    har & \cellcolor[HTML]{FFFFFF} 1 & \cellcolor[HTML]{78C2E0} 4 & \cellcolor[HTML]{D2EBF5} 2 & \cellcolor[HTML]{A5D6EA} 3 & \cellcolor[HTML]{4BAED6} 5\\ \hline
    hatespeech & \cellcolor[HTML]{FFFFFF} 1 & \cellcolor[HTML]{A5D6EA} 3 & \cellcolor[HTML]{D2EBF5} 2 & \cellcolor[HTML]{78C2E0} 4 & \cellcolor[HTML]{4BAED6} 5\\ \hline
    hiva & \cellcolor[HTML]{FFFFFF} 1 & \cellcolor[HTML]{A5D6EA} 3 & \cellcolor[HTML]{78C2E0} 4 & \cellcolor[HTML]{D2EBF5} 2 & \cellcolor[HTML]{4BAED6} 5\\ \hline
    imdb & \cellcolor[HTML]{FFFFFF} 1 & \cellcolor[HTML]{A5D6EA} 3 & \cellcolor[HTML]{78C2E0} 4 & \cellcolor[HTML]{4BAED6} 5 & \cellcolor[HTML]{D2EBF5} 2\\ \hline
    orl & \cellcolor[HTML]{FFFFFF} 1 & \cellcolor[HTML]{D2EBF5} 2 & \cellcolor[HTML]{A5D6EA} 3 & \cellcolor[HTML]{4BAED6} 5 & \cellcolor[HTML]{78C2E0} 4\\ \hline
    secom & \cellcolor[HTML]{FFFFFF} 1 & \cellcolor[HTML]{78C2E0} 4 & \cellcolor[HTML]{D2EBF5} 2 & \cellcolor[HTML]{A5D6EA} 3 & \cellcolor[HTML]{4BAED6} 5\\ \hline
    seismic & \cellcolor[HTML]{FFFFFF} 1 & \cellcolor[HTML]{A5D6EA} 3 & \cellcolor[HTML]{D2EBF5} 2 & \cellcolor[HTML]{78C2E0} 4 & \cellcolor[HTML]{4BAED6} 5\\ \hline
    sentiment & \cellcolor[HTML]{FFFFFF} 1 & \cellcolor[HTML]{A5D6EA} 3 & \cellcolor[HTML]{D2EBF5} 2 & \cellcolor[HTML]{78C2E0} 4 & \cellcolor[HTML]{4BAED6} 5\\ \hline
    sms & \cellcolor[HTML]{FFFFFF} 1 & \cellcolor[HTML]{D2EBF5} 2 & \cellcolor[HTML]{A5D6EA} 3 & \cellcolor[HTML]{4BAED6} 5 & \cellcolor[HTML]{78C2E0} 4\\ \hline
    svhn & \cellcolor[HTML]{FFFFFF} 1 & \cellcolor[HTML]{A5D6EA} 3 & \cellcolor[HTML]{78C2E0} 4 & \cellcolor[HTML]{D2EBF5} 2 & \cellcolor[HTML]{4BAED6} 5\\ \hline
    \end{tabularx}
    \vspace{0.15cm}
    \label{tab:rerun-table-NS}
    \vspace{-0.5cm}
\end{table}

\begin{table}[ht]
    \centering
    \caption{The order given by scale-normalized stress on embeddings of datasets (rows). Each embedding (MDS, t-SNE, UMAP, LLE, and ISO) is according to metric value, with lower ranks having lower stress.}
    \small
    \begin{tabularx}{\linewidth}{|p{1.6cm}| X X X X p{.9cm}|}
    \multicolumn{6}{c}{\textbf{SNS Ranking}}\\    
    \hline 
      & MDS & ISO & LLE & UMAP & t-SNE\\ \hline
    bank & \cellcolor[HTML]{FFFFFF} 1 & \cellcolor[HTML]{A5D6EA} 3 & \cellcolor[HTML]{4BAED6} 5 & \cellcolor[HTML]{78C2E0} 4 & \cellcolor[HTML]{D2EBF5} 2\\ \hline
    cifar10 & \cellcolor[HTML]{FFFFFF} 1 & \cellcolor[HTML]{D2EBF5} 2 & \cellcolor[HTML]{4BAED6} 5 & \cellcolor[HTML]{78C2E0} 4 & \cellcolor[HTML]{A5D6EA} 3\\ \hline
    cnae9 & \cellcolor[HTML]{FFFFFF} 1 & \cellcolor[HTML]{78C2E0} 4 & \cellcolor[HTML]{4BAED6} 5 & \cellcolor[HTML]{A5D6EA} 3 & \cellcolor[HTML]{D2EBF5} 2\\ \hline
    coil20 & \cellcolor[HTML]{FFFFFF} 1 & \cellcolor[HTML]{4BAED6} 5 & \cellcolor[HTML]{78C2E0} 4 & \cellcolor[HTML]{A5D6EA} 3 & \cellcolor[HTML]{D2EBF5} 2\\ \hline
    epileptic & \cellcolor[HTML]{FFFFFF} 1 & \cellcolor[HTML]{D2EBF5} 2 & \cellcolor[HTML]{78C2E0} 4 & \cellcolor[HTML]{4BAED6} 5 & \cellcolor[HTML]{A5D6EA} 3\\ \hline
    fashion\_mnist & \cellcolor[HTML]{FFFFFF} 1 & \cellcolor[HTML]{D2EBF5} 2 & \cellcolor[HTML]{4BAED6} 5 & \cellcolor[HTML]{78C2E0} 4 & \cellcolor[HTML]{A5D6EA} 3\\ \hline
    fmd & \cellcolor[HTML]{FFFFFF} 1 & \cellcolor[HTML]{A5D6EA} 3 & \cellcolor[HTML]{4BAED6} 5 & \cellcolor[HTML]{78C2E0} 4 & \cellcolor[HTML]{D2EBF5} 2\\ \hline
    har & \cellcolor[HTML]{FFFFFF} 1 & \cellcolor[HTML]{A5D6EA} 3 & \cellcolor[HTML]{4BAED6} 5 & \cellcolor[HTML]{78C2E0} 4 & \cellcolor[HTML]{D2EBF5} 2\\ \hline
    hatespeech & \cellcolor[HTML]{FFFFFF} 1 & \cellcolor[HTML]{A5D6EA} 3 & \cellcolor[HTML]{78C2E0} 4 & \cellcolor[HTML]{4BAED6} 5 & \cellcolor[HTML]{D2EBF5} 2\\ \hline
    hiva & \cellcolor[HTML]{FFFFFF} 1 & \cellcolor[HTML]{78C2E0} 4 & \cellcolor[HTML]{4BAED6} 5 & \cellcolor[HTML]{A5D6EA} 3 & \cellcolor[HTML]{D2EBF5} 2\\ \hline
    imdb & \cellcolor[HTML]{FFFFFF} 1 & \cellcolor[HTML]{4BAED6} 5 & \cellcolor[HTML]{78C2E0} 4 & \cellcolor[HTML]{A5D6EA} 3 & \cellcolor[HTML]{D2EBF5} 2\\ \hline
    orl & \cellcolor[HTML]{FFFFFF} 1 & \cellcolor[HTML]{78C2E0} 4 & \cellcolor[HTML]{4BAED6} 5 & \cellcolor[HTML]{D2EBF5} 2 & \cellcolor[HTML]{A5D6EA} 3\\ \hline
    secom & \cellcolor[HTML]{FFFFFF} 1 & \cellcolor[HTML]{78C2E0} 4 & \cellcolor[HTML]{4BAED6} 5 & \cellcolor[HTML]{A5D6EA} 3 & \cellcolor[HTML]{D2EBF5} 2\\ \hline
    seismic & \cellcolor[HTML]{FFFFFF} 1 & \cellcolor[HTML]{78C2E0} 4 & \cellcolor[HTML]{4BAED6} 5 & \cellcolor[HTML]{A5D6EA} 3 & \cellcolor[HTML]{D2EBF5} 2\\ \hline
    sentiment & \cellcolor[HTML]{FFFFFF} 1 & \cellcolor[HTML]{78C2E0} 4 & \cellcolor[HTML]{4BAED6} 5 & \cellcolor[HTML]{A5D6EA} 3 & \cellcolor[HTML]{D2EBF5} 2\\ \hline
    sms & \cellcolor[HTML]{FFFFFF} 1 & \cellcolor[HTML]{78C2E0} 4 & \cellcolor[HTML]{4BAED6} 5 & \cellcolor[HTML]{A5D6EA} 3 & \cellcolor[HTML]{D2EBF5} 2\\ \hline
    svhn & \cellcolor[HTML]{FFFFFF} 1 & \cellcolor[HTML]{D2EBF5} 2 & \cellcolor[HTML]{78C2E0} 4 & \cellcolor[HTML]{4BAED6} 5 & \cellcolor[HTML]{A5D6EA} 3\\ \hline

    \end{tabularx}
    \vspace{0.15cm}
    \label{tab:rerun-table-SNS}
    \vspace{-0.5cm}
\end{table}

\section{Discussion}

We showed the need to account for scale when using stress-based measures to validate DR results. Otherwise, the results can be misleading; for example, when using  normalized stress~\cite{DBLP:journals/tvcg/EspadotoMKHT21,DBLP:journals/corr/abs-2403-05882,DBLP:journals/bspc/DasanP21,DBLP:journals/corr/abs-2003-09017,DBLP:conf/sibgrapi/MarcilioEG17,DBLP:conf/apvis/AmorimBNSS14,DBLP:conf/ieeevast/AmorimBDJNS12,DBLP:journals/tvcg/PaulovichNML08,DBLP:journals/ivs/NevesFMFP18,DBLP:journals/cgf/PaulovichEPBMN11,DBLP:conf/eurova-ws/MachadoT023,DBLP:conf/igarss/ChenCG06,DBLP:conf/sigmod/FaloutsosL95,DBLP:journals/tvcg/NonatoA19}, it appears as though a  random, uncorrelated plot of the data does a better job at distance preservation than t-SNE (and sometimes even MDS). Intuitively, this should not be true, and visually, it clearly is not the case; see~\autoref{fig:teaser}. 

Using a metric called ``normalized stress" seems like a good idea when evaluating the quality of a new DR method. However, as we have shown in this paper, ``normalized stress" is very similar to ``raw stress," and both of them are severely affected by scaling. 

We argue that using such metrics is not fair when comparing the outputs from different DR methods, which often come at very different scales. 
The scale-invariant metrics are not only more in line with the expected results; they are more consistent and cannot be easily manipulated (e.g., by selectively scaling some outputs). Non-metric stress and Shepard goodness score have the benefit of an easy-to-interpret range, while scale-normalized stress is a simple computation added onto the already widely adopted normalized stress. For this reason, we prefer to recommend scale-normalized stress for use in further DR evaluation, but any of the three scale-invariant metrics are preferable over the commonly used scale-sensitive measures.






\section{Limitations, Conclusions and Future Work}

\subsection{Limitations}
While the experiments described here provide insights into the behavior of different stress metrics used to evaluate  DR techniques, we acknowledge several limitations.

We base our experimental validation on selecting three (\hyperref[sec:experiment-A]{Experiment A}) or five (\hyperref[sec:experiment-B]{Experiment B}) DR techniques and five metrics (raw stress, normalized stress, Shepard goodness, non-metric stress, and scale-normalized stress).
The experiments utilized a fairly small number of datasets. While we believe this provides a reasonable base for comparison, the results may not generalize to different datasets or different embedding techniques. 
Including more DR techniques, more datasets, and more stress-based metrics~\cite{DBLP:conf/vissym/SeifertSK10,ed0a397fae134f218fa4fd3998ced318,DBLP:journals/tvcg/KorenC04} would increase the robustness of our findings.

In \hyperref[sec:experiment-A]{Experiment A}, we operated under the expectation that there is a “correct order” when considering how well three DR techniques preserve distances in their embeddings. However plausible the hypothesis that random embeddings should score worse than embeddings that explicitly attempt to preserve distances in the computed embeddings, a better-grounded experiment could be designed to verify the effectiveness of stress-based metrics.

Finally,
\hyperref[sec:experiment-A]{Experiment A}
does not account for potential variations in the performance of DR techniques due to hyperparameter settings or initialization conditions, which can influence the quality of DR techniques~\cite{DBLP:conf/vluds/EngelHH11}. A future analysis should test the DR techniques under a variety of conditions to assess their performance more comprehensively.

\subsection{Conclusions and Future Work}
In this paper, our analysis of normalized stress demonstrates its significant susceptibility to scale. Our empirical evidence reveals that the scale-sensitivity problem affects real-world datasets, and taking scale into account may alter previous evaluations.
To address this, we considered scale-invariant distance-based quality metrics and also proposed a solution to the scale sensitivity of the most popular normalized stress method, scale-normalized stress. 
 
Although our results call into question the validity of evaluations made using scale-sensitive stress metrics, it still remains for future work to revisit other prior studies that used these metrics and verify their results. There are other scale-sensitive DR quality metrics, such as KL divergence~\cite{DBLP:conf/nips/HintonR02}. It seems prudent and worthwhile to consider the scale-invariance of such DR quality metrics to improve the embedding techniques themselves as well as their evaluation.

\newpage
\bibliographystyle{abbrv-doi-hyperref}

\bibliography{references}

\end{document}